\documentclass[runningheads]{llncs}
\usepackage[utf8]{inputenc}
\usepackage[T5]{fontenc} 
\usepackage{tabularx} 
\usepackage{float}
\usepackage{subcaption}
\usepackage{amsmath, amssymb, amsfonts}
\usepackage{graphicx}
\usepackage{enumitem}
\usepackage{footmisc}
\usepackage{booktabs}
\usepackage{makecell}

\usepackage{multirow} 
\usepackage{bbm}
\usepackage[none]{hyphenat}
\usepackage[final]{microtype}
\usepackage{marvosym}
\usepackage{float}

\usepackage[font=small,skip=2pt]{caption}
\AtBeginDocument{%
  \setlength{\abovedisplayskip}{6pt}
  \setlength{\belowdisplayskip}{6pt}
}

\usepackage[colorlinks=true,linkcolor=blue,urlcolor=blue,citecolor=blue]{hyperref}

\begin{document}
\title{ViConBERT: Context-Gloss Aligned Vietnamese Word Embedding for Polysemous and Sense-Aware Representations}
%
%
\author{Khang T. Huynh\inst{1,2,3},
Dung H. Nguyen\inst{1,2},
Binh T. Nguyen\inst{1,2, }\textsuperscript{\Letter}
}
\authorrunning{Khang T. Huynh et al.}
%
\institute{University of Information Technology, Ho Chi Minh City, Vietnam
\email{23560020@gm.uit.edu.vn, dungngh@uit.edu.vn, binhnt@uit.edu.vn}
\and
Vietnam National University, Ho Chi Minh City, Vietnam
\and
Birmingham City University, Birmingham, UK}
\titlerunning{ViConBERT}
\maketitle              

\begin{abstract}
\sloppy

Recent advances in contextualized word embeddings have greatly improved semantic tasks such as Word Sense Disambiguation (WSD) and contextual similarity, but most progress has been limited to high-resource languages like English. Vietnamese, in contrast, still lacks robust models and evaluation resources for fine-grained semantic understanding. In this paper, we present \textbf{ViConBERT}, a novel framework for learning Vietnamese contextualized embeddings that integrates contrastive learning (SimCLR) and gloss-based distillation to better capture word meaning. We also introduce \textbf{ViConWSD}, the first large-scale synthetic dataset for evaluating semantic understanding in Vietnamese, covering both WSD and contextual similarity. Experimental results show that ViConBERT outperforms strong baselines on WSD (F1 = 0.87) and achieves competitive performance on ViCon (AP = 0.88) and ViSim-400 (Spearman’s $\rho$ = 0.60), demonstrating its effectiveness in modeling both discrete senses and graded semantic relations. Our code, models, and data are available at \url{https://github.com/tkhangg0910/ViConBERT}.

\end{abstract}

\section{Introduction}

Understanding word meaning in context is central to tasks such as word sense disambiguation (WSD) and contextual semantic similarity. It remains a fundamental challenge in NLP with broad applications in machine translation, question answering, and information retrieval. Accurate contextual modeling enables systems to distinguish between multiple senses of a word and capture subtle semantic nuances crucial for downstream tasks.

While substantial progress has been made in high-resource languages like English with transformer-based models~\cite{devlin2019bert}, advances for low-resource languages such as Vietnamese remain limited. Fine-grained semantic understanding in Vietnamese faces two main gaps: (i) the absence of embedding models explicitly trained for sense-level discrimination, and (ii) the scarcity of publicly available benchmark datasets for contextual word semantic evaluation.

Recent approaches integrating contextual embeddings with contrastive learning show promise for capturing semantic distinctions but are mostly developed for high-resource settings. For Vietnamese, available datasets are either small, inaccessible~\cite{vicon,vsimlex}, or not tailored for WSD and contextual word similarity~\cite{xnlivn,stsbvn}, leaving the field without a standardized basis for progress.

To bridge this gap, we introduce a novel framework for learning contextualized Vietnamese word embeddings via word–gloss distillation. In addition, we construct ViConWSD, a new synthetic benchmark dataset tailored for evaluating fine-grained semantic understanding in Vietnamese. By systematically generating sense-labeled examples from gloss definitions, our dataset provides large-scale supervision where manual annotations are infeasible.
Our key contributions are:

\begin{enumerate}
\item We propose a training pipeline that leverages gloss alignment to enhance contextual embeddings in low-resource settings, with a focus on Vietnamese.
\item We curate and release ViConWSD, the first large-scale synthetic benchmark for Vietnamese WSD and contextual similarity.
\item We conduct extensive experiments showing consistent improvements over strong baselines across multiple semantic tasks.
\end{enumerate}

\section{Related Works}

Transformer-based models like BERT and RoBERTa~\cite{devlin2019bert,roberta} achieve strong results on semantic tasks via context-sensitive embeddings. Multilingual variants (XLM-R~\cite{xlmr}) and Vietnamese-specific models (PhoBERT, ViDeBERTa~\cite{phobert,videberta}) extend these advances to Vietnamese. Yet their generic objectives capture broad context but struggle with fine-grained semantics.

To address such challenges, WSD and semantic similarity have been extensively studied. Early WSD approaches relied on lexical resources like WordNet~\cite{wordnet}, with evaluation datasets such as SemCor and SemEval~\cite{semeval2007}. Recent neural methods~\cite{glossbert} integrate gloss definitions (PolyBERT, BEM~\cite{polybert,bem}), with contrastive learning increasingly used to align contextual and gloss embeddings~\cite{simclr}. Similarly, semantic similarity has progressed from static embeddings~\cite{word2vec,fasttext} to contextualized approaches like SimCSE~\cite{simcse}. Standard benchmarks include SimLex-999~\cite{simlex} and WiC~\cite{wic}.

Despite this progress, most methods and resources focus on high-resource languages. Vietnamese NLP still faces a scarcity of large-scale semantic datasets: resources like Vietnamese WordNet support syntax and tagging, while semantic datasets such as ViCon and VSimLex-999~\cite{vicon,vsimlex} remain limited in size and coverage. This motivates our work to introduce a scalable semantic framework for Vietnamese, combining contrastive learning with gloss embeddings and providing richer evaluation resources for fine-grained semantic understanding.

\section{Our Proposed ViConBERT Model}

\begin{figure}[H]
\includegraphics[width=\textwidth]{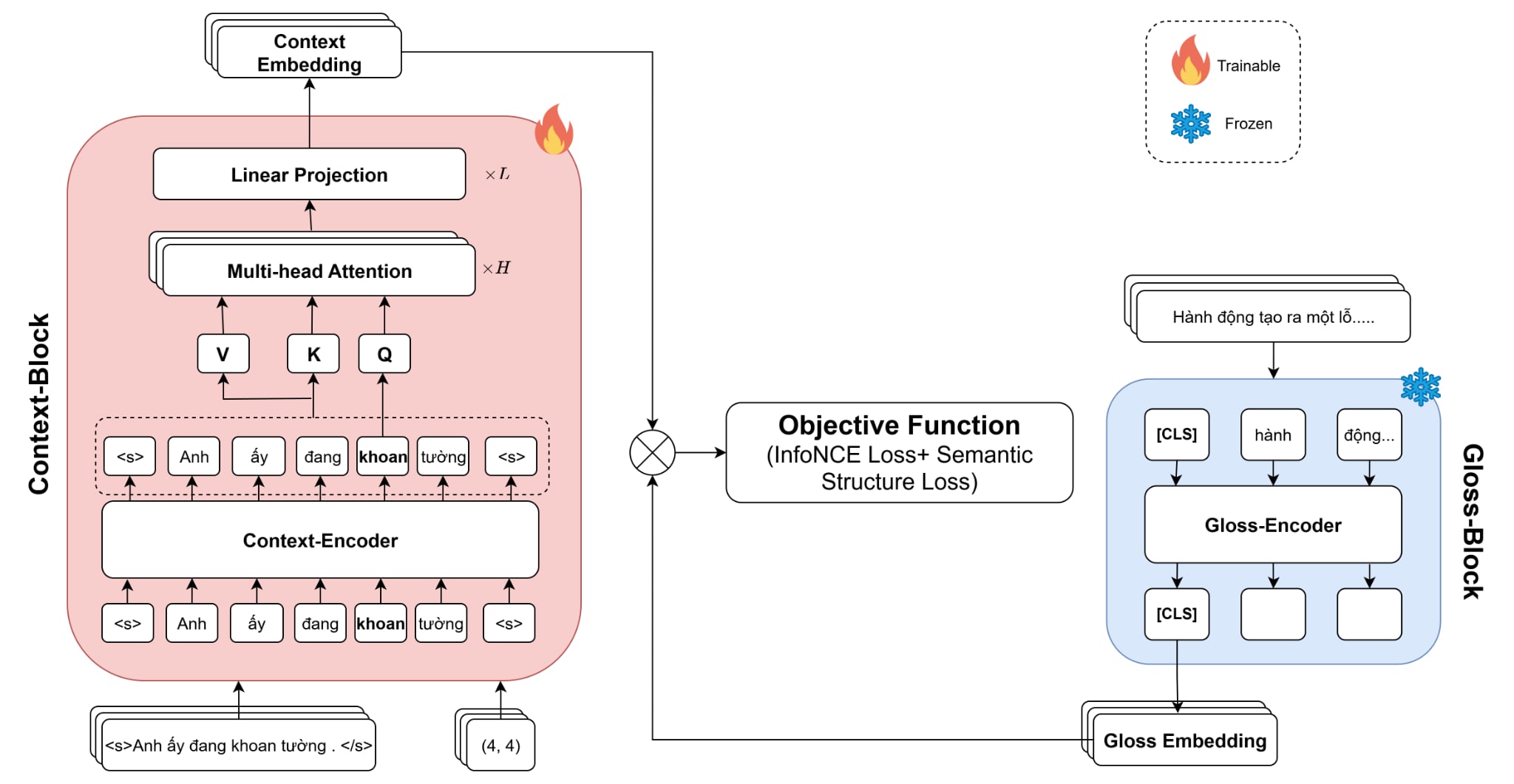}
\caption{Training architecture of ViConBERT. Left (red): the Context Encoder processes sentences (e.g., "Anh ấy đang khoan tường." He is drilling the wall) with the target word khoan (drill) via multi-head attention and projection to produce contextual embeddings. Right (blue): the Gloss Encoder encodes glosses (e.g., "Hành động tạo ra một lỗ..." An action that creates a hole...) into gloss embeddings. The objective combines InfoNCE to align context and gloss and Semantic Structure Loss to preserve their relative semantic structure. Context block is trainable; gloss block is frozen.} \label{fig1}
\end{figure}

We propose \textbf{ViConBERT} (Vietnamese-Context BERT), a model designed to represent the semantic space of contextualized word embeddings. The overall architecture is illustrated in Figure~\ref{fig1}. 

The key idea behind ViConBERT is to \textit{distill} the gloss embedding space of a well-pretrained sentence embedding model into a context encoder through a contrastive learning framework~\cite{simclr}. Specifically, the model aligns the contextual representation of a target word with the vector representation of its corresponding gloss (definition).

Unlike previous gloss-based architectures such as \textbf{GlossBERT}, \textbf{BEM}, and \textbf{PolyBERT} that treat word sense disambiguation as a discrete classification task, \textbf{ViConBERT} models sense semantics in a \textit{continuous embedding space}. This allows the model to capture graded similarity between related senses rather than assigning isolated sense labels. Moreover, by distilling the gloss space from a pretrained sentence embedding model, ViConBERT obtains stable semantic anchors for each word, making the training more efficient and less dependent on complex negative sampling strategies.

Let $C$ be a tokenized context sentence containing the target word $W$ at token indices $I$. The context block $B_c$ produces a contextual embedding $E_{C} = B_c(C, I)$. Similarly, let $G$ denote the gloss corresponding to $W$. The gloss block $B_g$ generates the gloss embedding $E_{G} = B_G(G)$. We use $C_{B}$ and $G_{B}$ to denote the matrices containing all $E_{C}$ and $E_{G}$ vectors in a mini-batch.

\subsection{Context Block And Gloss Block}

\subsubsection{Context Block.} The context encoder $f_c$ is a pretrained BERT-based model that processes input tokens $C = [T^{1}, \dots, T^{n}]$ to produce contextual embeddings $H_{C} = f_c(C) \in \mathbb{R}^{n \times d_h}$. We extract embeddings at the target word span $I = (start\_pos, end\_pos)$ and apply mean pooling to obtain query vector $Q \in \mathbb{R}^{1 \times d_h}$. Using multi-head attention with $Q$ as query and $H_C$ as key-value, we compute:

\begin{equation}
\Tilde{H_c} = \text{MultiHeadAttn}(Q, H_C, H_C)
\end{equation}

\noindent Finally, we project to the joint embedding space:

\begin{equation}
E_C= \Tilde{H_c}  \cdot W_C
\end{equation}

\noindent where $W_C \in \mathbb{R}^{d_h \times d_{model}}$.

\subsubsection{Gloss Block.} We employ a pretrained sentence embedding model optimized for semantic textual similarity as the gloss encoder $f_g$:

\begin{equation}
E_G= f_g(G)
\end{equation}

\subsection{Objective Function}

\subsubsection{Gloss-Context Alignment Loss} ($\mathcal{L}_{\text{InfoNCE}}$) encourages context embeddings to align with corresponding gloss embeddings using InfoNCE loss:

\begin{equation}
\mathcal{L}_{\text{InfoNCE}} = \frac{1}{N} \sum_{i=1}^{N} -\log \left( \frac{\sum_{j \in P_i} \exp(g_i^\top c_j/\tau)}
                  {\sum_{k \neq i} \exp(g_i^\top c_k/\tau) + \epsilon} \right)
\end{equation}

\noindent where $\tau$ is temperature and $P_i$ denotes positive pairs for sample $i$.

\subsubsection{Semantic Structure Loss} ($\mathcal{L}_{\text{SS}}$) is our proposed loss that promotes structural consistency between the context and gloss embedding spaces. Specifically, it encourages the relative distances (i.e., semantic topology) among context embeddings to mirror those among their corresponding gloss embeddings. We compute dissimilarity matrices:

\begin{equation}
D_C = 1 - C_BC_B^\top \quad \text{and} \quad D_G = 1 - G_BG_B^\top
\end{equation}

\noindent where $C_B =[E_C^{(1)},...,E_C^{(N)}]$ and $G_B =[E_G^{(1)},...,E_G^{(N)}]$ represent the contextual and gloss embedding matrices in a batch.

The Semantic Structure Loss is computed as the mean squared error (MSE) between these dissimilarity matrices:

\begin{equation}
\mathcal{L}_{\text{SS}} = \frac{1}{N^2} \sum_{i=1}^{N} \sum_{j=1}^{N} \left( D_C^{(i,j)} - D_G^{(i,j)} \right)^2
= \operatorname{MSE}(D_C, D_G)
\end{equation}

The final training objective combines both losses:

\begin{equation}
\mathcal{L}_{\text{Total}} = \mathcal{L}_{\text{InfoNCE}} + \lambda \mathcal{L}_{\text{SS}}
\end{equation}

\noindent where $\lambda$ balances the two loss terms.

\section{Performance Evaluation of ViConBERT}
\subsection{ViConWSD Dataset Creation}

\begin{figure}
\includegraphics[width=\textwidth]{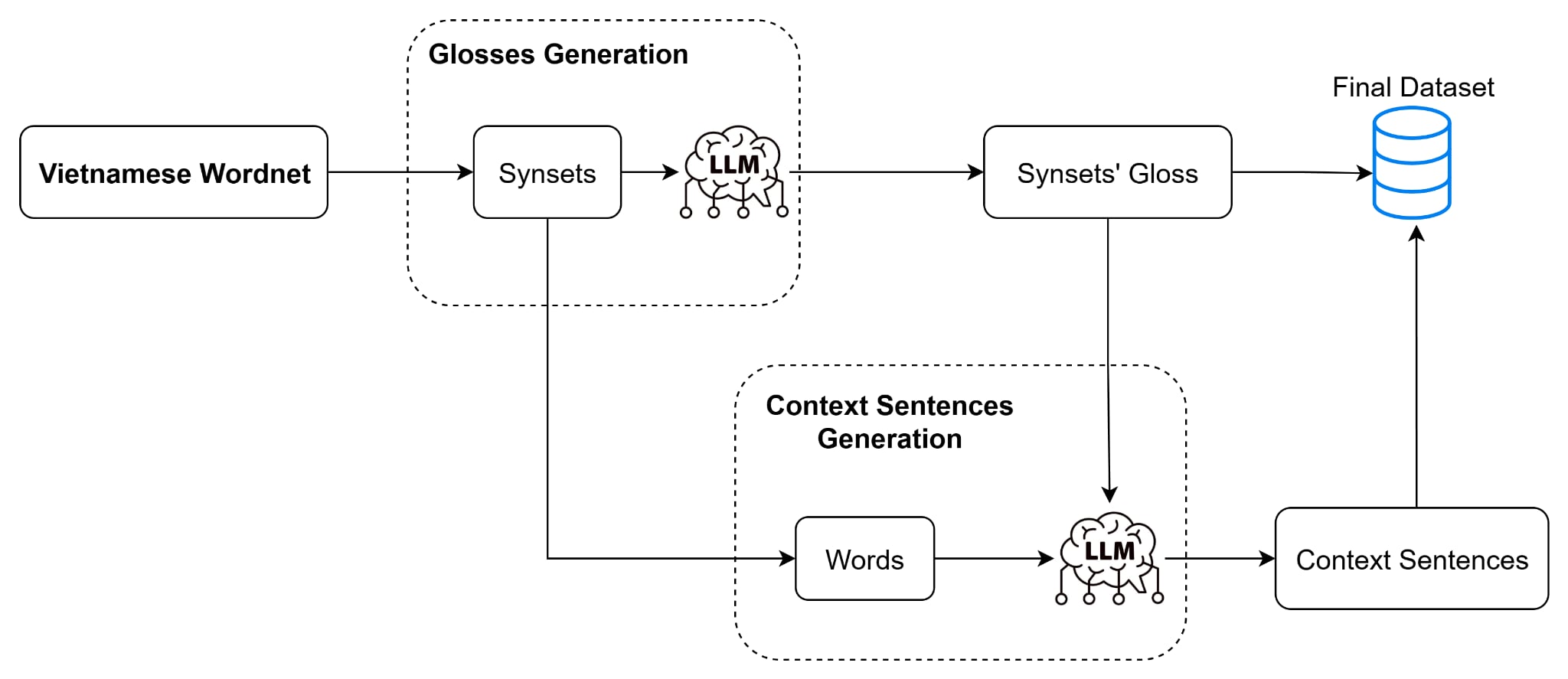}
\caption{Synthetic dataset construction pipeline.} \label{dataset_pipeline}
\end{figure}

We present the first synthetic Vietnamese dataset designed for semantic understanding tasks, including word sense disambiguation and semantic similarity, by leveraging gloss and supersense information.

The dataset is derived from Vietnamese WordNet\footref{note:vwordnet} and consists of synsets - sets of synonymous words categorized into supersenses, i.e., coarse-grained semantic classes that group related senses into broader categories.

\textit{Construction Pipeline:} The dataset is automatically generated through a multi-stage pipeline (Figure~\ref{dataset_pipeline}), inspired by prior semantic datasets but optimized for low-resource languages and minimal manual annotation:

\begin{enumerate}
    \item \textbf{Vietnamese WordNet:} We use the publicly available Vietnamese WordNet corpus\footnote[1]{\label{note:vwordnet}\url{https://github.com/zeloru/vietnamese-wordnet}}, which aggregates synsets from public sources such as websites and Wikipedia.
    
    \item \textbf{Gloss Generation:} LLMs trained on Vietnamese (specifically \texttt{Gemini 2.5}\footnote[2]{\url{https://ai.google.dev/gemini-api/docs/}}) are prompted under strict constraints to produce fine-grained glosses (definitions) for each synset. A single gloss is shared across the synonymous words in the synset.

    \item \textbf{Context Sentence Generation:} Conditioned on both the target word and its gloss, LLMs (including \texttt{LLaMA~3.3~70B}\footnote[3]{\url{https://huggingface.co/meta-llama/Llama-3.3-70B-Instruct}}, \texttt{Qwen3~32B}\footnote[4]{\url{https://huggingface.co/Qwen/Qwen2.5-32B-Instruct}}, and \texttt{DeepSeek-R1-Distill-LLaMA~70B}\footnote[5]{\url{https://huggingface.co/deepseek-ai/DeepSeek-R1-Distill-Llama-70B}}) are employed to generate context-rich sentences that demonstrate word usage while maintaining semantic alignment with the synset.

\end{enumerate}

This synthetic pipeline produces the final dataset, whose key statistics and characteristics are summarized in Table~\ref{dataset_statistics}.

\begin{table}[ht]
\centering
\begin{tabular}{|l|c|}
\hline
\textbf{Statistics} & \textbf{Value} \\
\hline
Number of synsets & {33471} \\
\hline
Number of words & {100160} \\
\hline
Average words per synset & {3} \\
\hline
Average sentences per word & {22.8} \\
\hline
Total Number of Polysemous and Homonym & {5292(5\% of dataset)}  \\
\hline
\end{tabular}
\caption{Comprehensive statistics of our dataset.}
\label{dataset_statistics}
\end{table}

\paragraph{Quality Verification.}
To assess the semantic quality of the synthetic ViConWSD dataset, we randomly sampled 200 gloss–context pairs and manually examined their coherence. Approximately 90\% of the pairs were judged to be semantically consistent and contextually appropriate, suggesting that the automatic generation pipeline produces data of sufficiently high quality for downstream training.

\subsection{Training and Hyperparameter Configurations}
All experiments were conducted on a single NVIDIA RTX A6000 GPU, with models trained for 100 epochs. The largest model required about 60 hours.  

\subsubsection{Context Block.}
We use PhoBERT \cite{phobert} as the context encoder, truncating or padding inputs to its 256-token limit. Encoder outputs are passed through a multi-head attention layer and a final linear projection to obtain context embeddings. Hyperparameters are shown in Table~\ref{tab:context_hyperparams}.

\begin{table}[ht]
\centering
\begin{tabular}{|l|c|c|}
\hline
\textbf{Hyperparameter} & \textbf{ViConBERT-base} & \textbf{ViConBERT-large} \\
\hline
Pretrained model & PhoBERT-base & PhoBERT-large \\
Embedding dimension & 768 & 1024 \\
Attention heads & 3 & 4 \\
Projection layers & 1 & 2 \\
Projection output dim. & 768 & 768 \\
Dropout rate & 0.3 & 0.3 \\
Parameters & 140M & 376M \\
\hline
\end{tabular}
\caption{Context block hyperparameters.}
\label{tab:context_hyperparams}
\end{table}

\subsubsection{Gloss Block.} 
We adopt \texttt{dangvantuan/vietnamese-embedding}\footnote[6]{\label{note:dv}\url{https://huggingface.co/dangvantuan/vietnamese-embedding}} as the gloss encoder, which achieves state-of-the-art performance on the Vietnamese STS task.

\subsubsection{Training Configurations.}
We train the context encoder (\textit{base}) and task-specific layers (\textit{custom}) with separate learning rates of $4\times10^{-5}$ and $3\times10^{-4}$ to balance pretrained knowledge retention and task adaptation. Optimization uses AdamW with a \texttt{ReduceLROnPlateau} scheduler. Models are trained for 100 epochs with a batch size of 768. The InfoNCE loss employs a temperature $\tau = 0.3$, and the Semantic Structure loss has an auxiliary weight $\lambda = 1$.

\subsection{Model Evaluation and Metrics}
We conducted three sets of experiments:
\begin{enumerate}
   \item \textbf{PLM fine-tuning:} Comparing different PLMs as gloss and context encoders.  
   \item \textbf{Comparison with state-of-the-art WSD models:} Benchmarking against strong baselines.  
   \item \textbf{Evaluation on standard semantic benchmarks:} Testing on widely used Vietnamese benchmarks.  
\end{enumerate}

\subsubsection{Pretrained language model fine-tuning.}

We fine-tune various PLMs - XLM-RoBERTa~\cite{xlmr}, PhoBERT~\cite{phobert}, and ViDeBERTa~\cite{videberta} - as gloss and context encoders in a retrieval-based setup. Performance is measured by \textbf{F1@k} and \textbf{NDCG@k}. The gloss encoders are \texttt{dangvantuan/vietnamese-embedding} (DV) and \texttt{VoVanPhuc/sup-SimCSE-VietNamese-phobert-base} (VP).

\begin{table}[ht]
\renewcommand{\arraystretch}{1.4}
\centering
\resizebox{0.95\textwidth}{!}{
\begin{tabular}{|c|c|c|c|c|c|c|c|}
\hline
\multicolumn{2}{|c|}{\textbf{Model}} & \multicolumn{3}{c|}{\textbf{F1@k (\%)}} & \multicolumn{3}{c|}{\textbf{NDCG@k (\%)}} \\
\hline
\textbf{Gloss encoder} & \textbf{Context encoder} 
 & k=1 & k=5 & k=10 & k=1 & k=5 & k=10 \\
\hline
\multirow{5}{*}{DV} 
  & $XLMR_{Base}$        & 76.73 & 81.56 & 81.56 & 81.76 & 86.86 & 87.56 \\
\cline{2-8}
  & $XLMR_{Large}$       & \textbf{77.44} & \textbf{81.77} &83.15 & \textbf{82.33} & 86.77 & 87.47 \\
\cline{2-8}
  & $PhoBERT_{Base}$     & 76.87 & 81.61 & \textbf{83.18} & 81.97 & \textbf{86.94} & \textbf{87.63} \\
\cline{2-8}
  & $PhoBERT_{Large}$    & 76.52 &81.51 & 83.11 & 81.68 & 86.83 & 87.53 \\
\cline{2-8}
  & $ViDeBERTa_{Base}$  & 72.36 & 78.36 &80.53 & 76.65 & 84.21 & 85.14 \\
\hline
\multirow{5}{*}{VP}
  & $XLMR_{Base}$        & 76.46 & 81.29 & 82.86 & \textbf{82.86} & 86.49 & 87.21 \\
\cline{2-8}
  & $XLMR_{Large}$       & \textbf{77.52}  & 81.56 & 82.91 & 82.32 & 86.49 & 87.19 \\
\cline{2-8}
  & $PhoBERT_{Base}$     & 76.63 & \textbf{81.59} & \textbf{83.20} & 81.77 & \textbf{86.92} & \textbf{87.62} \\
\cline{2-8}
  & $PhoBERT_{Large}$    & 76.42 & 81.45 & 83.13 & 81.41 & 86.68 & 87.41 \\
\cline{2-8}
  & $ViDeBERTa_{Base}$  & 72.52 & 78.60 & 80.79 & 76.16 & 83.82 & 84.81 \\
\hline
\end{tabular}}
\caption{Evaluation results of ViConBERT variants.}
\label{tab:model_comparison_final}
\end{table}

Table~\ref{tab:model_comparison_final} shows that larger encoders generally improve performance, with $PhoBERT_{Base}$ achieving the best overall results (83.18 F1@10, 87.63 NDCG@10). ViDeBERTa$_{Base}$ lags significantly behind. The performance gap between DV and VP gloss encoders is minimal, suggesting that context encoder choice is more critical than gloss encoder selection.

\subsubsection{Comparison with state-of-the-art WSD models.}
We evaluate WSD on a benchmark containing 5,292 words and 6,856 glosses, focusing on polysemous and homonymous cases. We compare ViConBERT against BEM~\cite{bem} and PolyBERT~\cite{polybert} using $PhoBERT_{base}$ as the shared encoder.

\begin{table}[ht]
\centering
\begin{tabularx}{\textwidth}{|c|c|>{\centering\arraybackslash}X|>{\centering\arraybackslash}X|>{\centering\arraybackslash}X|>{\centering\arraybackslash}X|}
\hline
\textbf{Model} & \textbf{Test Dataset} & \multicolumn{4}{c|}{\textbf{Different POS of Test Datasets}} \\
\cline{3-6}
 &  & Nouns & Verbs & Adj. & Adv. \\
\hline
BEM                 & 0.84 & 0.81 & \textbf{0.89} & 0.82 & 0.79 \\
\hline
PolyBERT            & 0.83 & 0.81 & \textbf{0.89} & 0.77  & 0.73 \\
\hline
\textbf{ViConBERT (Ours)} & \textbf{0.87} & \textbf{0.87} & 0.88 & \textbf{0.87} & \textbf{0.84} \\
\hline
\end{tabularx}
\caption{Performance of ViConBERT on WSD task compared with previous works.}
\label{tab:wsd_comparison}
\end{table}

Table~\ref{tab:wsd_comparison} shows that ViConBERT achieves the highest overall F1-score (0.87), notably outperforming BEM (0.84) and PolyBERT (0.83) on nouns and adjectives where lexical ambiguity is more prevalent. While BEM and PolyBERT slightly surpass ViConBERT on verbs (0.89 vs. 0.88), our model delivers more balanced performance across all parts of speech.

\subsubsection{Evaluation on standard semantic benchmarks.}

We test ViConBERT on two Vietnamese datasets: \textbf{ViCon}~\cite{vicon} (synonym–antonym pairs, evaluated using AP) and \textbf{ViSim-400}~\cite{vsimlex} (400 word pairs, evaluated using Spearman's $\rho$).

\begin{table}[ht]
\centering
{
\begin{tabular}{|c|c|ccc|}
\hline
\textbf{Model} & \textbf{ViSim-400} \\
\hline
SGNS     & 0.37   \\
\hline
mLCM        & 0.60 \\
\hline
dLCE        & \textbf{0.62} \\
\hline
\textbf{ViConBERT(Ours)}     & 0.60 \\
\hline
\end{tabular}}
\caption{Spearman's correlation $\rho$ on ViSim-400.}
\label{tab:visim}
\end{table}

\begin{table}[ht]
\centering
\begin{tabularx}{\textwidth}{|c|
    >{\centering\arraybackslash}X|
    >{\centering\arraybackslash}X|
    >{\centering\arraybackslash}X|
    >{\centering\arraybackslash}X|
    >{\centering\arraybackslash}X|
    >{\centering\arraybackslash}X|}
\hline
\textbf{Model} & \multicolumn{2}{c|}{\textbf{ADJ}} & \multicolumn{2}{c|}{\textbf{NOUN}} & \multicolumn{2}{c|}{\textbf{VERB}}  \\
\cline{2-7}
 & SYN & ANT & SYN & ANT & SYN & ANT \\
\hline
PPMI               & 0.70 & 0.38 & 0.68 & 0.39 & 0.69 & 0.38 \\
\hline
PLMI               & 0.59 & 0.44 & 0.61 & 0.42 & 0.63 & 0.41 \\
\hline
$weight^{SA}$      & \textbf{0.93} & \textbf{0.31} & \textbf{0.94} & \textbf{0.31} & \textbf{0.96} & \textbf{0.31}  \\
\hline
PPMI + SVD         & 0.76 & 0.36 & 0.66 & 0.40 & 0.81 & 0.34 \\
\hline
PLMI + SVD         & 0.49 & 0.51 & 0.55 & 0.46 & 0.51 & 0.49 \\
\hline
$weight^{SA}$+SVD  & 0.91 & 0.32 & 0.81 & 0.34 & 0.92 & 0.32 \\
\hline
\textbf{ViConBERT(Ours)} & 0.88 & 0.34 & 0.81 & 0.37 & 0.84 & 0.35 \\
\hline
\end{tabularx}
\caption{AP evaluation on ViCon.}
\label{tab:vicon}
\end{table}

As shown in Table~\ref{tab:visim} and Table~\ref{tab:vicon}, on ViSim-400, ViConBERT achieves 0.60, on par with mLCM and close to dLCE (0.62). On ViCon, ViConBERT demonstrates competitive synonym detection (0.88 ADJ, 0.81 NOUN, 0.84 VERB) and  antonym detection across all categories (0.34 ADJ, 0.37 NOUN, 0.35 VERB). These results highlight ViConBERT's robustness in handling both similarity and contrastive relations.

Unlike Weight\textsuperscript{SA}, which is optimized for synonym–antonym discrimination, ViConBERT learns a continuous contextual space instead of a binary relational boundary. Its InfoNCE loss separates senses without enforcing antonym repulsion, and ViConWSD focuses on gloss alignment rather than antonym supervision. Consequently, ViConBERT yields more balanced results on SYN and ANT, reflecting broader semantic generalization.

\subsection{Embedding Space Visualization}
We analyze the contextualized word representations for three cases:

\subsubsection{Homonym words.} 
\begin{figure}[htbp]
    \centering
    \begin{subfigure}[b]{0.33\textwidth}
        \includegraphics[width=\textwidth]{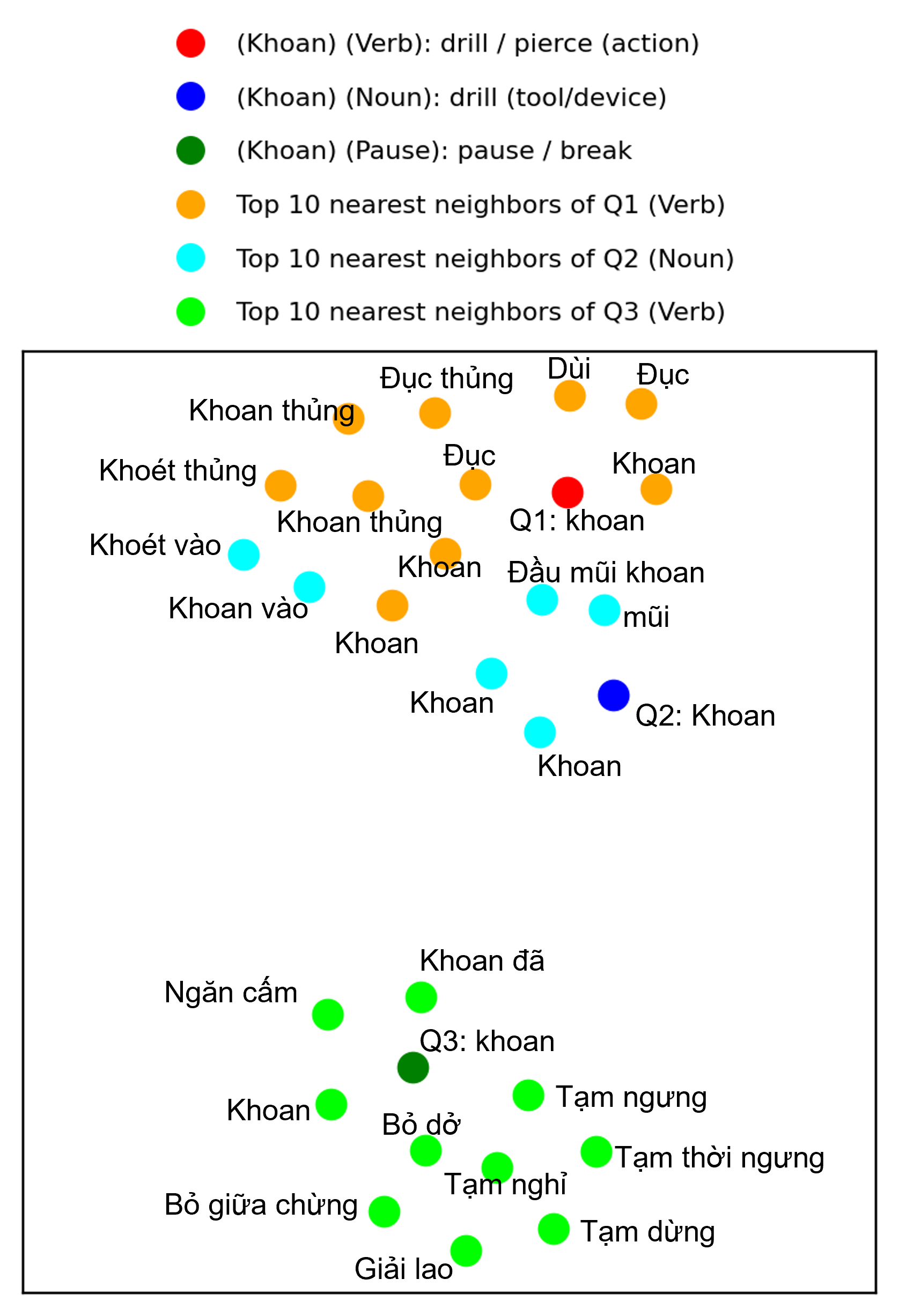}
        \caption{Homonym}
        \label{fig:hono}
    \end{subfigure}
    \hfill
    \begin{subfigure}[b]{0.32\textwidth}
        \includegraphics[width=\textwidth]{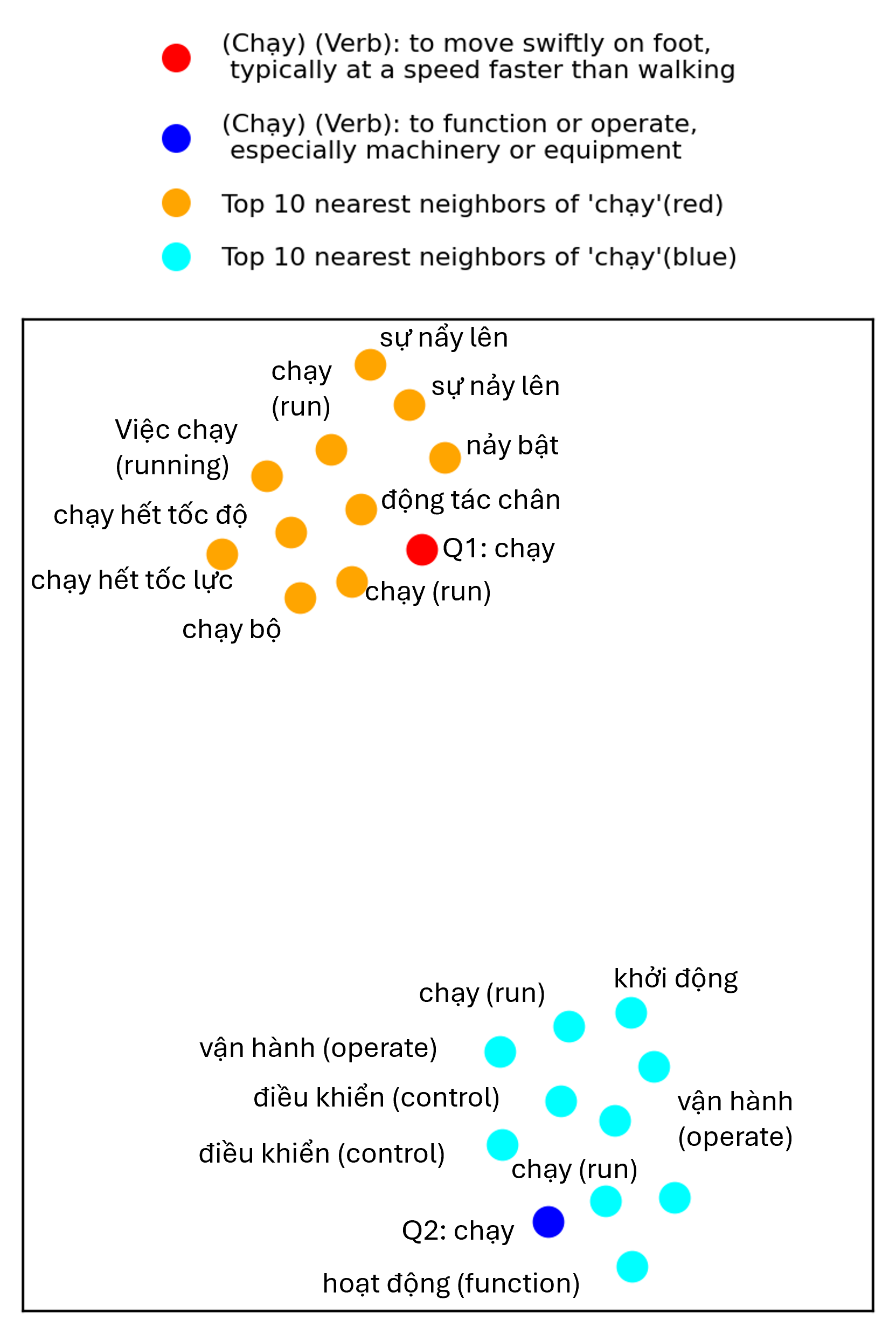}
        \caption{Polysemy}
        \label{fig:poly}
    \end{subfigure}
    \hfill
    \begin{subfigure}[b]{0.33\textwidth}
        \includegraphics[width=\textwidth]{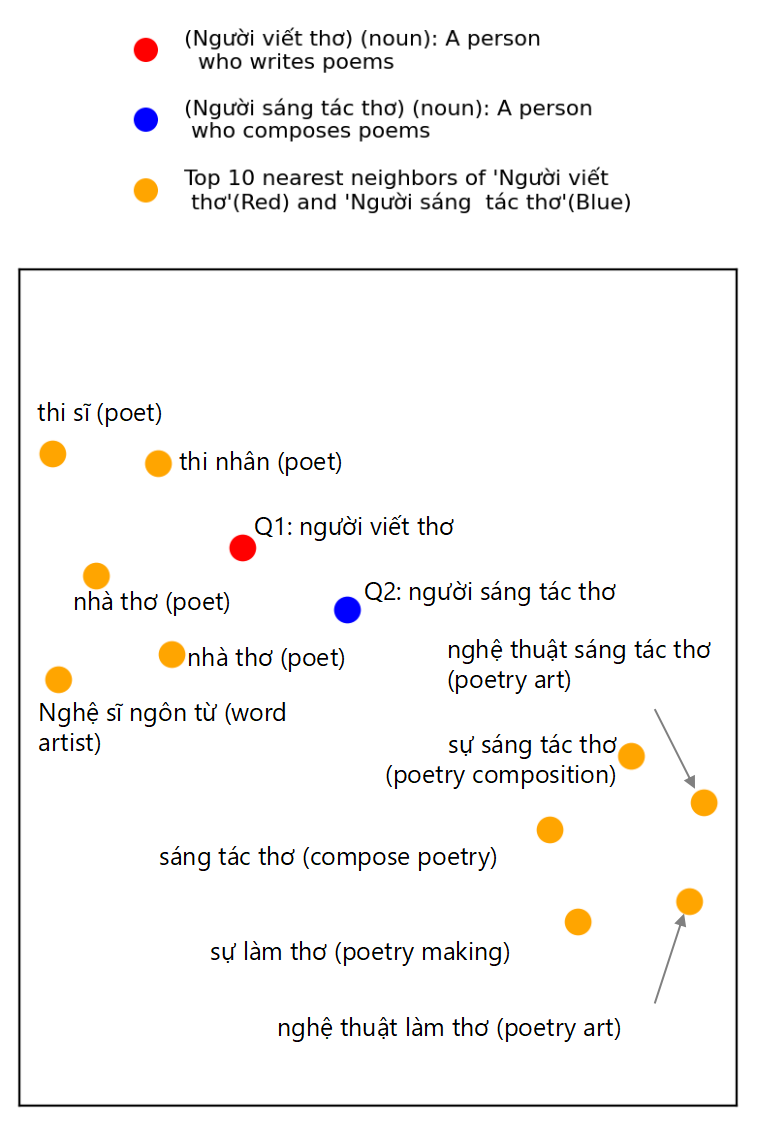}
        \caption{Zero-shot}
        \label{fig:zero}
    \end{subfigure}
    \caption{Embedding space for different word types}
    \label{fig:all_pipelines}
\end{figure}

We visualize the embedding space of "khoan," which has three distinct meanings: (1) drilling action, (2) drill tool, and (3) "stop/pause." Figure~\ref{fig:hono} shows how the model separates these senses with related words: (1) đục (bore), khoét (carve); (2) cái khoan, mũi khoan (drill bit); (3) khoan đã (hold up), tạm dừng (pause).

\subsubsection{Polysemous words}
We visualize "chạy" (run), which has two main senses: (1) moving swiftly on foot, and (2) functioning/operating. Figure~\ref{fig:poly} shows clear separation between these senses.

\subsubsection{Zero-shot (unseen) words}
We assess generalization to unseen terms: "người viết thơ" (person who writes poems) and "người sáng tác thơ" (person who composes poems), which are specific forms of "thi sĩ" or "nhà thơ" (poet). Figure~\ref{fig:zero} illustrates the model's ability to generalize to unseen concepts.

The visualizations confirm that ViConBERT effectively distinguishes different semantic senses across homonyms, polysemous words, and unseen expressions, with clear separation while preserving semantic relatedness within clusters.

\section{Conclusion and Future Work}
We proposed ViConBERT, a Vietnamese contextualized embedding model for capturing fine-grained semantic distinctions, along with a corresponding dataset. Experiments on retrieval tasks, WSD benchmarks, and semantic similarity datasets demonstrate consistent improvements over state-of-the-art baselines. Embedding visualizations show well-separated word senses and effective generalization to unseen terms.

Our study has limitations. First, although we manually assessed 200 randomly sampled pairs, overall dataset cleanliness cannot be fully guaranteed: rare words may have been exposed to large LLMs, and the WordNet-based resource is not official. Second, Vietnam lacks extensive WSD benchmarks, limiting our evaluation to our constructed dataset. Nevertheless, modern Vietnamese LLMs, trained on massive multilingual corpora, have captured a broad range of Vietnamese words and senses, providing a sufficiently reliable basis for evaluation.

Future work involves enhancing dataset quality through expert validation, cross-validating with different LLMs, and developing additional Vietnamese benchmarks for more comprehensive evaluation across domains and tasks.

\section*{Acknowledgement}

This research was supported by The VNUHCM-University of Information Technology's Scientific Research Support Fund.

\bibliographystyle{splncs04}
\bibliography{bibliography}

\end{document}